\pdfoutput=1

\documentclass[11pt]{article}

\usepackage[final]{acl}

\usepackage{times}
\usepackage{latexsym}
\usepackage{amsmath, amssymb}
\usepackage{hyperref}
\usepackage{graphicx}
\usepackage{cite}
\usepackage{geometry}
\usepackage{booktabs}
\usepackage{multirow} 
\usepackage{array}
\usepackage{makecell}   
\usepackage{subcaption} 
\usepackage{float} 
\usepackage[T1]{fontenc}
\usepackage{placeins}

\usepackage[utf8]{inputenc}

\usepackage{microtype}

\usepackage{graphicx}

\title{DTCRS: Dynamic Tree Construction for Recursive Summarization}

\author{
  Guanran Luo\thanks{Equal contribution.}, 
  Zhongquan Jian\footnotemark[1], 
  Wentao Qiu ,
  {\bf Meihong Wang, Qingqiang Wu} \\
   School of Informatics, Xiamen University \\
  \texttt{luoguanran@stu.xmu.edu.cn, wuqq@xmu.edu.cn}
}

\newcolumntype{C}[1]{>{\centering\arraybackslash}m{#1}}

\begin{document}
\maketitle
\begin{abstract}
Retrieval-Augmented Generation (RAG) mitigates the hallucination issues of large language models (LLMs) by integrating external knowledge. For abstractive questions involving multi-step reasoning, knowledge from multiple sections is often required. To address this issue, recent research has introduced recursive summarization, which constructs a hierarchical summary tree by clustering text chunks, integrating information from various parts of the document to provide evidence for abstractive questions. However, summary trees often contain a large number of redundant summary nodes, which not only increase construction time but may also negatively impact question answering. Moreover, recursive summarization is not suitable for all types of questions. We introduce DTCRS, a method that dynamically generates summary trees based on document structure and query semantics. DTCRS determines whether a summary tree is necessary by analyzing the question type. It then decomposes the question and uses the embeddings of sub-questions as initial cluster centers, reducing redundant summaries while improving the relevance between summaries and the question. Our approach significantly reduces summary tree construction time and achieves substantial improvements across three QA tasks. Additionally, we investigate the applicability of recursive summarization to different question types, providing valuable insights for future research.
\end{abstract}

\section{Introduction}

Although Large Language Models (LLMs) have demonstrated tremendous advantages across various tasks, updating model knowledge to adapt to the ever-changing world remains a critical issue \citep{zhang2023large}. Compared to methods of fine-tuning models, the Retrieval-Augmented Generation (RAG) paradigm, which combines LLMs with knowledge bases by injecting external knowledge, can update knowledge without modifying model parameters \citep{wang2024knowledge}. RAG has become an important method for mitigating large model hallucinations and enhancing answer interpretability \citep{li2024enhancing}. 

\begin{figure}
  \centering
  \includegraphics[width=\linewidth]{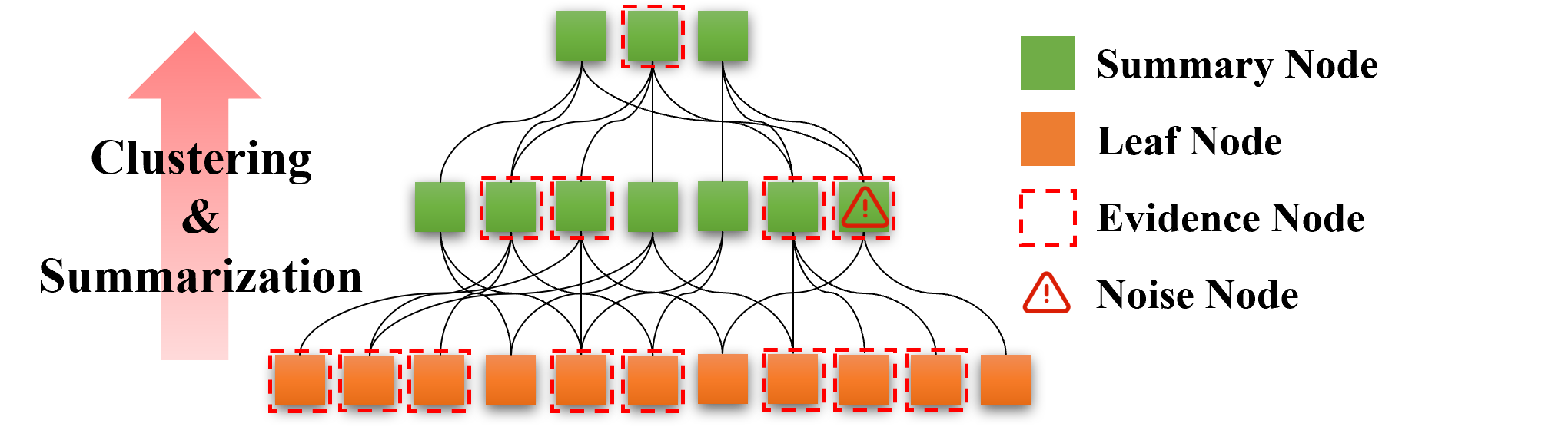}
  \caption{The presence of numerous query-irrelevant summary nodes in the summary tree increases construction time and may adversely affect correct answers.}
  \label{fig:motivation}
\end{figure}

Most current RAG research only retrieves consecutive short text chunks, which lack sufficient context to answer questions that require integrating information from multiple parts of a document. Recursive summarization \citep{wang2023recursively} addresses this issue by hierarchically clustering and summarizing dispersed text chunks to construct a summary tree, thereby integrating the scattered information. Wu et al. propose a method for recursively summarizing entire fiction novels using human feedback \citep{wu2021recursively}. HIBRIDS introduces hierarchical biases into the Transformer architecture to better capture document structure for long-document summarization \citep{cao2022hibrids}. RAPTOR \citep{sarthi2024raptor} establishes a new baseline for RAG by constructing hierarchical summary trees for information retrieval, outperforming existing retrieval methods across various QA datasets.

Recursive summarization is effective because it generates summaries at different granularity levels, providing more information for answer generation. However, as shown in Figure~\ref{fig:motivation}, we find that it also introduces a large amount of redundant summaries unrelated to the answer, which not only increases computational overhead but may also negatively impact the correctness of the final answer. One key reason for this issue is that traditional summary tree is a static tree based on documents, which splits the original document into text chunks and then generates summaries based on clustered chunks \citep{sarthi2024raptor}. As a result, the summary tree only reflects the document itself rather than capturing the semantics of the query, leading to an abundance of redundant summaries that are unrelated to the query.

Moreover, whether all types of questions can benefit from recursive summarization remains an unresolved issue. Intuitively, recursive summarization integrates dispersed knowledge within a document, making it particularly beneficial for abstractive questions that require multi-step reasoning. However, for simpler tasks such as extractive or boolean questions, whether recursive summarization provides any advantage remains uncertain \citep{zhang2022extractive}. Therefore, indiscriminately using recursive summarization is likely to interfere with the generation of answers.

To address the above issues, we propose a Dynamic Tree Construction for Recursive Summarization (DTCRS), which replaces the static summary tree generated solely from the document with a dynamically constructed summary tree based on the document structure and query semantics. This approach enhances the relevance between the summary topics and the query while reducing redundant summaries. Specifically, we first determine the question type. For complex questions that require summarizing information from multiple parts of the document, we generate a table of contents (ToC) for the document. Based on this ToC, we decompose the question into multiple simpler sub-questions. DTCRS then uses the number of sub-questions and their embeddings as the number of clusters and initial cluster centers to perform Gaussian Mixture Model (GMM) clustering on text chunks and generate corresponding text summaries. This process is repeated recursively to construct a summary tree for RAG.

We conduct comprehensive experiments on three QA datasets, and the results show significant improvements over the baselines, indicating that the dynamic summary tree generates more relevant summaries for the questions. We further enhance answer quality by assessing question types before building the summary tree, thereby reducing time overhead. Additionally, we analyze the performance of DTCRA across various types of questions. The experimental results validate our hypothesis that introducing recursive summarization significantly enhances the performance of LLMs on more challenging abstractive questions. However, for extractive and boolean questions, recursive summarization provides no advantage and may even have a negative impact.

\section{Related Work}

Effectively utilizing long-range context remains a persistent challenge in retrieval-augmented QA. Dai et al. \citep{dai2019transformer} introduced Transformer-XL to overcome fixed-length context limitations; however, recent studies by Sun et al. \citep{sun2021long} and Liu et al. \citep{liu2024lost} highlight that contemporary language models still encounter difficulties in fully leveraging extended contexts. To address these challenges, research has explored several complementary directions, which we group here into four major areas.

\textbf{Retrieval-Augmented Generation.}
RAG improves QA performance by incorporating external knowledge into model inference. Techniques such as self-reflective prompting, adaptive complexity control, and iterative refinement have proven effective in improving query formulation \citep{asai2023self, jeong2024adaptive, chan2024rq}. These approaches tend to work well within narrow domains but face difficulties generalizing to more complex or diverse inputs \citep{zhang2024raft, siriwardhana2023improving}. By introducing a summarization structure that absorbs and organizes retrieved content, our method aims to maintain the advantages of RAG while reducing its reliance on brittle query-retrieval dependencies.

\textbf{Retrieval Methods.}
Open-domain QA has benefited significantly from improvements in retrieval, especially with the rise of dense representations and neural scoring. Early works based on term frequency \citep{sparck1972statistical} have evolved into embedding-based retrieval \citep{khattab2020colbert, guu2020retrieval, karpukhin2020dense, liu2021dense} that offers better contextual alignment. More recent methods adopt hierarchical or multi-stage retrieval schemes \citep{arivazhagan2023hybrid}, and knowledge distillation has helped balance efficiency and effectiveness \citep{izacard2020distilling, roberts2020much, min2021joint}. While most of these techniques aim to improve recall and ranking quality, they are typically designed as independent modules. In contrast, our strategy connects retrieval output directly to a downstream structure, improving continuity and interpretability.

\textbf{Text Summarization.}
Summarization plays a key role in condensing long or multi-document inputs. Recent approaches have explored recursive summarization to extend memory \citep{wang2023recursively} and query-focused methods to improve relevance \citep{deng2023nonfactoid, xu2020coarse}. Chunk-level and weakly supervised summarization \citep{angelidis2018summarizing, gao2023enabling, sarthi2024raptor, wu2021recursively} further increase scalability. These techniques primarily focus on fluency or informativeness, often at the expense of structural clarity. By organizing summaries within a tree-based layout aligned to document sections, we aim to retain coherence while supporting targeted reasoning.

\textbf{Query Enhancement.}
Improving the clarity and utility of user questions has become a central technique in QA pipelines. Query rewriting has been shown to improve retrieval for under-specified or ambiguous queries \citep{mo2024aligning, peng2024large}, while decomposition supports multi-hop inference \citep{reppert2023iterated, ye2023large, radhakrishnan2023question}. These strategies typically operate as pre-processing steps. In contrast, our approach grounds sub-question generation in the structure of the Summary Tree, using the natural organization of documents to inform and constrain the reasoning process.

Together, these lines of work provide a strong foundation. By drawing on their strengths and embedding them within a unified, structured representation, our method aims to enhance long-context QA in a more interpretable and modular fashion.

\section{DTCRS}
\begin{figure*}[t]
  \centering
\includegraphics[width=\textwidth]{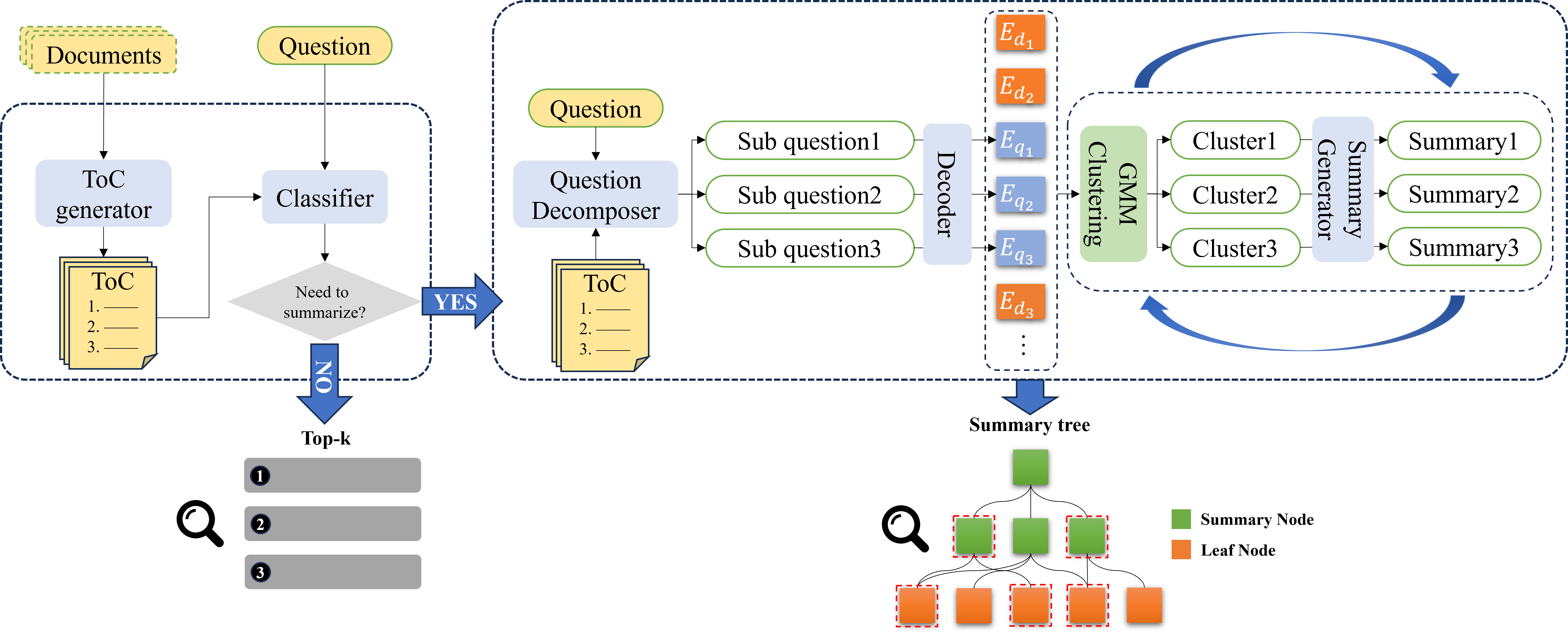}  
  \caption{The overall process of DTCRS. The classifier first determines the question type. If summarization of document information is required, a dynamic summary tree is constructed; otherwise, DPR \citep{karpukhin2020dense} is used for retrieval. During the construction of the dynamic summary tree, the question is decomposed based on the document table of contents, with sub-questions serving as initial cluster centers for recursive clustering to generate summaries at different levels.}
  \label{fig:framework}
\end{figure*}

The overall process of DTCRS is shown in Figure~\ref{fig:framework}. In this section, we first introduce how to classify the questions, then explain how DTCRS handles different types of questions, with a focus on how DTCRS constructs dynamic recursive summary trees for abstractive questions. Finally, we introduce two retrieval methods for the summary tree.

\subsection{Question Type Classification}

Indiscriminately introducing recursive summarization adds extra overhead and may negatively impact the performance of question answering. Therefore, it is necessary to determine the question type before constructing the recursive tree in order to decide whether recursive summarization is required. We use LLM as a table of contents (ToC) generator and classifier. The ToC \(c\) is generated based on the document \(d\), and then \(c\) and the original question \(q\) are input into the classifier. It outputs a binary label:
\begin{equation}
y = f_{\text{LLM}}(q, c) \in \{0,1\}
\end{equation}
The classification criteria are whether the question is complex and whether it requires summarizing information from multiple sections of the ToC to provide a complete answer. If the classifier returns \(1\), a dynamic summary tree is constructed for that question; otherwise, we use the DPR\citep{karpukhin2020dense} method to retrieve the top \(k\) text chunks:
\begin{equation}
S(d, q) = \text{Top}_k \left( \{ \text{sim}(q, T_i) \mid T_i \in d \} \right)
\end{equation}
Where \( T_i \) represents the \( i \)-th text chunk in document \( d \). \( \text{sim}(q, T_i) \) computes the similarity between query \( q \) and text chunk \( T_i \).

We provide the prompts for table of contents generation and question classification in Appendix~\ref{sec:appendix1}.

\subsection{Dynamic Summary Trees}
Our summary trees are dynamic because they generate different summary trees for different questions. There are two main issues that dynamic summary trees address: (1) the redundancy of nodes in the summary tree, and (2) the irrelevance of generated summaries to the query. To address these issues, we describe the construction process of dynamic summary trees below.

\paragraph{Quesion Decomposition.}To address the redundancy issue in summary trees, we can reduce the number of clusters at each layer. Previous methods have employed the Bayesian Information Criterion (BIC) \citep{neath2012bayesian} to determine the optimal number of clusters for the model:
\begin{equation}
\text{BIC} = \ln(n) \ p - 2 \ln(\hat{L}),
\end{equation}
where \( n \) is the number of data points, \( p \) is the number of parameters in the model, and \( \hat{L} \) is the maximized value of the likelihood function of the model. 

The number of clusters is determined by the number of text chunks and the number of model parameters. Therefore, the longer the text and the more model parameters, the more redundant nodes may be generated. Assume that the document \( d \) is divided into \( N_d \) chunks, the total computational effort can be described by the geometric series:
\begin{equation}
N_d + \frac{N_d}{2} + \frac{N_d}{4} + \frac{N_d}{8} + \cdots
\end{equation}
The total computational workload is \(2N_d\). In contrast, we use an LLM as the question decomposer and input both the ToC \(c\) and the original question \(q\) into the question decomposer to generate a set of sub-questions \(Q' = \{ q_1, q_2, \dots, q_j \}\).

There are two reasons for introducing the ToC: one is to limit the scope of the sub-questions to the themes of the reference document, and the second is to better align the sub-questions with the different sections of the document. 

Then, we use \(Q'\) for the first layer of clustering, so the number of sub-questions \(N_{Q'}\) corresponds to the number of clusters, and the embeddings of the sub-questions \(E_{Q'}\) can serve as the initial cluster centers, resulting in the total computational effort:
\begin{equation}
N_d + N_{Q'} + \frac{N_{Q'}}{2} + \cdots
\end{equation}
Since \(N_{Q'} \ll N_d\), the computational effort can be considered approximately \(N_d\), thereby reducing the time required for clustering and generating summaries. Moreover, by reducing the number of clusters, there is no need to employ hierarchical clustering to capture the relationships from themes to details among texts \citep{sarthi2024raptor}, thus further enhancing the efficiency of clustering.

We provide prompts for decomposing questions in Appendix~\ref{sec:appendix1}.

\paragraph{Text Chunk Segmentation.} The complexity of the summary tree scales linearly with the length of the document \citep{sarthi2024raptor}. Therefore, although there are methods available for segmenting text chunks based on semantics \citep{sawarkar2024blended}, given that constructing the summary tree itself incurs significant time overhead, if semantic segmentation is used, then for a dataset such as NarrativeQA \citep{kovcisky2018narrativeqa}—where a document can exceed 100,000 tokens—the time required to build a summary tree would be intolerable. Therefore, we use a fixed segmentation method, and we set the text chunk size to 500 tokens. Any sentence that extends beyond the 500-token boundary is moved in its entirety to the next text chunk to avoid having an incomplete sentence within a single text chunk.

\paragraph{Clustering.} To address the issue of generated summaries being potentially irrelevant to the query, we use the number of sub-questions as the number of clusters, with embeddings of sub-questions serving as initial clustering centers. This quesion-oriented clustering approach enhances the relevance of summaries to the questions.

We use a specific encoder to convert text chunks into embeddings, and then, following previous research \citep{sarthi2024raptor}, we employ Gaussian Mixture Models (GMM) for soft clustering, allowing a text chunk to be assigned to multiple categories. GMM assumes that all data points are generated from a finite number of Gaussian distributions, each corresponding to a cluster. The probability density function of the entire model is given by:
\begin{equation}
p(x|\theta) = \sum_{m=1}^M \pi_m \, \mathcal{N}(x|\mu_m, \Sigma_m),
\end{equation}
where \( \pi_m \) is the mixing weight of the \( m \)-th Gaussian distribution, satisfying \(\textstyle 0 \leq \pi_m \leq 1\) and \(\textstyle \sum_{m=1}^M \pi_m = 1\), and \( \theta \) represents the set of all parameters to be estimated, including all mixing weights, means, and covariance matrices.

To better measure the similarity between embeddings \citep{aggarwal2001surprising}, we use Uniform Manifold Approximation and Projection (UMAP) for dimensionality reduction \citep{mcinnes2018umap}. First, we combine all text chunk embeddings \(E_T = \{ e_{t_1}, e_{t_2}, \dots, e_{t_p} \}\). and sub-question embeddings \(E_{Q'} = \{ e_{q_1}, e_{q_2}, \dots, e_{q_j} \}\) into a new embedding set, then perform unified dimensionality reduction to ensure semantic consistency between sub-questions and text chunks. Formally, this dimensionality reduction process can be expressed as:
\begin{equation}
E_{\text {reduced }}=\Phi_{\mathrm{UMAP}}\left(E_T \oplus E_{Q'}\right)
\end{equation}

We use global clustering instead of the hierarchical clustering algorithm \citep{sarthi2024raptor} because, on one hand, the number of sub-questions is small and hierarchical clustering typically requires a larger number of clusters; on the other hand, global clustering is more efficient. In the first layer of clustering, we use the number of sub-questions as the number of clusters, with embeddings serving as the initial clustering centers. This naturally sets two adaptive hyperparameters for clustering. After the first layer, we do not consider sub-questions but use the number of clusters determined by BIC and random initial clustering centers, as the number of sub-questions by then is greater than or equal to the number of text chunks.

\paragraph{Recursive Summarization Generation.}The clustered text chunks are fed into an LLM-based summarization generator to produce respective summaries. This process is then repeated by reducing dimensions and clustering again until further clustering is no longer possible. Although the generated summaries may contain slight hallucinations, these hallucinations do not significantly impact the question-and-answer results \citep{sarthi2024raptor,zhang2022extractive}. The prompts used for generating summaries are provided in Appendix~\ref{sec:appendix1}.

\subsection{Retrieval}
For the two retrieval methods of the summary tree: tree traversal and collapsed tree \citep{sarthi2024raptor}, the tree traversal method selects the top \( k \) most relevant nodes based on cosine similarity at each layer, considers the children of these selected nodes in the next layer, and then chooses the \( k \) nodes with the highest cosine similarity to the query vector, repeating this process until reaching the leaf nodes. The collapsed tree method unfolds all nodes as a single layer, selecting the top \( k \) nodes with the highest cosine similarity score to the query, and continuously adds nodes to the result set until a predefined maximum token count is reached. Since the performance of the collapsed tree consistently outperforms tree traversal \citep{sarthi2024raptor}, we use the collapsed tree method for retrieval in subsequent experiments.

\section{Experimental Setup}
\subsection{Settings}
We use GPT-4 \citep{achiam2023gpt}, GPT-4o-mini, and DeepSeek-V2-Lite-Chat (7B) \citep{liu2024deepseek} as the LLMs, which are among the most advanced closed-source and open-source LLMs currently available. The selected LLMs are used for question classification, ToC generation, sub-question generation, summarization, and question answering. The embedding model is SBERT \citep{reimers2019sentence}, with the GMM clustering threshold set to 0.5. The maximum length for summaries is 100 tokens, the maximum length for text chunks is 500 tokens. During retrieval, DPR's top-k is set to 5, the maximum token limit for the collapsed tree is set to 3500, and FAISS is used for nearest neighbor search \citep{johnson2019billion}. The experiments are conducted on an NVIDIA A800 80GB PCIe GPU and an Intel(R) Xeon(R) Silver 4314 CPU @ 2.40GHz.

\subsection{Datasets}
\begin{table}[h]
\centering
\resizebox{\columnwidth}{!}{%
\begin{tabular}{lcc}
\toprule
\textbf{Dataset} & \textbf{Number of Questions} & \textbf{Average Document Characters} \\
\midrule
QASPER & 1451 & 21,889.87 \\
QuALITY & 2128 & 24,723.06 \\
NarrativeQA & 10,558 & 332,133.92 \\
\bottomrule
\end{tabular}
}
\caption{Dataset Overview.}
\label{tab:dataset_overview}
\end{table}

\begin{table}[h]
\centering
\footnotesize 
\begin{tabular}{lc}
\toprule
\textbf{Type} & \textbf{Number} \\
\midrule
Extractive & 501 (34.53\%) \\
Abstractive & 513 (35.35\%) \\
Boolean & 239 (16.47\%) \\
Unanswerable & 198 (13.65\%) \\
\bottomrule
\end{tabular}
\caption{Statistics of Different Types of Questions in QASPER.}
\label{tab:qasper_question_types}
\end{table}

Our experiments utilize three question-answering datasets: NarrativeQA \citep{kovcisky2018narrativeqa}, QASPER \citep{dasigi2021dataset}, and QuALITY \citep{pang2021quality}. The basic statistics of these datasets are presented in Table\ref{tab:dataset_overview} and Table\ref{tab:qasper_question_types}. These datasets enable the evaluation of DTCRS's performance across various document lengths and diverse question types.

\paragraph{NarrativeQA:} This dataset contains reference documents paired with unique question-answer pairs. During evaluation, the model generates answers based on the provided document and question, which are then compared directly to the unique ground-truth answers using standard metrics such as BLEU-1, BLEU-4, ROUGE-L, and METEOR.

\paragraph{QASPER:} This dataset comprises shorter documents with multiple answers per question provided by different annotators. Questions are categorized into answerable/unanswerable, yes/no, abstractive, and extractive types. Evaluation is conducted by calculating the token-level F1 score between the model's prediction and all provided answers, selecting the highest score as the final evaluation metric.

\paragraph{QuALITY:} QuALITY includes context paragraphs paired with multiple-choice questions. During evaluation, the model selects the most appropriate answer from provided options based on the context and question. The chosen answers are then compared against the gold standard answers to measure performance through accuracy.

\subsection{Comparisons}
\paragraph{DPR \citep{karpukhin2020dense}.}Maps queries and documents into a dense vector space and finds the most relevant documents by computing the similarity between vectors.
\paragraph{RAPTOR \citep{sarthi2024raptor}.} Recursively summarizes text in a hierarchical manner, constructing a tree structure that allows the model to retrieve information from different levels of abstraction during inference. This method captures both local details and the overall structure and themes of documents, thereby better supporting complex reasoning tasks.
\paragraph{State-of-the-Art Methods on Each Task.} \textbf{NarrativeQA}: BiDAF \citep{seo2016bidirectional}, BM25 + BERT\citep{robertson2009probabilistic,devlin2018bert}, Recursively Summarizing Books \citep{wu2021recursively}, Retriever + Reader \citep{izacard2020distilling}. \textbf{QuALITY}: Longformer-base \citep{beltagy2020longformer}, DPR + DeBERTaV3-large \citep{karpukhin2020dense,he2020deberta}, CoLISA + DeBERTaV3-large \citep{dong2023colisa}. \textbf{QASPER}: LongT5 XL \citep{guo2021longt5}, CoLT5 XL \citep{ainslie2023colt5}.

\section{Experimental Results}
\subsection{Main Results}
\label{sec:main_results}
\begin{table}[h]
\centering
\footnotesize 
\begin{tabular}{lc}
\toprule
\textbf{Model} & \multicolumn{1}{c}{\textbf{F1}} \\
\midrule
DPR + GPT-4 & \multicolumn{1}{c}{51.3} \\
LongT5 XL & \multicolumn{1}{c}{53.1} \\
CoLT5 XL & \multicolumn{1}{c}{53.9} \\
RAPTOR + GPT-4 & \multicolumn{1}{c}{55.7} \\
\textbf{DTCRS + GPT-4} & \multicolumn{1}{c}{\textbf{58.5}} \\
\bottomrule
\end{tabular}
\caption{Experimental results on QASPER test set using GPT-4 as the language model. The results of other methods are reported as stated in their original papers. DTCRS outperforms RAPTOR and achieves the best results.}
\label{tab:qasper_results}
\end{table}
\begin{table}[h]
\centering
\resizebox{\columnwidth}{!}{%
\begin{tabular}{lcccc}
\toprule
\multirow{2}{*}{\makecell[c]{\textbf{Model}}} & \multicolumn{2}{c}{\textbf{Accuracy}} & \multicolumn{2}{c}{\textbf{SAT-style Score}} \\
\cmidrule(lr){2-3} \cmidrule(lr){4-5}
 & \textbf{Test Set} & \textbf{Hard Subset} & \textbf{Test Set} & \textbf{Hard Subset} \\
\midrule
Longformer-base & 30.7 & 29.3 & 7.6 & 5.7 \\
DPR + DeBERTaV3-large & 55.4 & 46.1 & 40.5 & 28.1 \\
CoLISA + DeBERTaV3-large & 62.3 & 54.7 & 49.7 & 39.6 \\
DPR + GPT-4o-mini & 62.8 & 50.3 & 52 & 35.9 \\
RAPTOR + GPT-4o-mini & 67.6 & 57 & 59 & 44.9 \\
\textbf{DTCRS + GPT-4o-mini} & \textbf{74.7} & \textbf{62.9} & \textbf{63.7} & \textbf{48.2} \\
\bottomrule
\end{tabular}
}
\caption{Experimental results on QuALITY test set using GPT-4o-mini as the language model. The results of other methods are reported as stated in their original papers. DTCRS achieves the best performance on both the entire dataset and the challenging subset.}
\label{tab:quality_results}
\end{table}
\begin{table}[h]
\centering
\resizebox{\columnwidth}{!}{%
\begin{tabular}{lcccc}
\toprule
\textbf{Model} & \textbf{ROUGE-L} & \textbf{BLEU-1} & \textbf{BLEU-4} & \textbf{METEOR} \\
\midrule
BiDAF & 6.2 & 5.7 & 0.3 & 3.7 \\
BM25 + BERT & 15.5 & 14.5 & 1.4 & 5.0 \\
Recursively Summarizing Books & 21.6 & 22.3 & 4.2 & 10.6 \\
DPR + GPT-4o-mini & 26.5 & 21.5 & 1.3 & 14.2 \\
RAPTOR + GPT-4o-mini & 25.0 & 21.2 & 1.1 & 14.1 \\
Retriever + Reader & \textbf{32.0} & \textbf{35.3} & \textbf{7.5} & 11.1 \\
\textbf{DTCRS + GPT-4o-mini} & 26.0 & 21.1 & 1.3 & \textbf{14.4} \\
\bottomrule
\end{tabular}
}
\caption{Experimental results on NarrativeQA test set using GPT-4o-mini as the language model. The results of other methods are reported as stated in their original papers. DTCRS achieves the best METEOR score but does not show an overall advantage compared to DPR and RAPTOR using the same language model. We attribute this to the lack of complex questions requiring summarization and reasoning in NarrativeQA.}
\label{tab:NarrativeQA_results}
\end{table}

We compare DTCRS with state-of-the-art methods on each task, and the experimental results are the best values from two runs on the test set. As shown in Tables~\ref{tab:qasper_results} and~\ref{tab:quality_results}. DTCRS+GPT-4 achieves the highest F1 score of 58.5\% on the QASPER dataset, while DTCRS+GPT-4o-mini attains the highest accuracy of 74.7\% and 62.9\% on the full and hard subsets of the QuALITY dataset, respectively. This demonstrates that the recursive summaries generated by our method provide useful references for LLMs in answering questions.

As shown in Table~\ref{tab:NarrativeQA_results}, DTCRS+GPT-4o-mini achieves the highest METEOR score of 14.4\% on the NarrativeQA dataset but performs worse than the Retriever+Reader \citep{izacard2020distilling} method on other metrics. We attribute the varying performance of DTCRS across different tasks to differences in question types and difficulty. QASPER contains a significant proportion of abstractive questions that require synthesizing scattered information, while QuALITY includes a challenging hard subset where DTCRS exhibits a more pronounced advantage. Although the NarrativeQA dataset provides an ultra-long document, its questions are predominantly simple extractive questions, such as “Who is Eve?” and “Where does Ralston recover?”. These types of questions do not benefit from DTCRS, which explains why DTCRS performs worse than Retriever+Reader and exhibits similar performance to DPR and RAPTOR on this dataset.

\begin{figure}[h]
    \centering
    \includegraphics[scale=0.35]{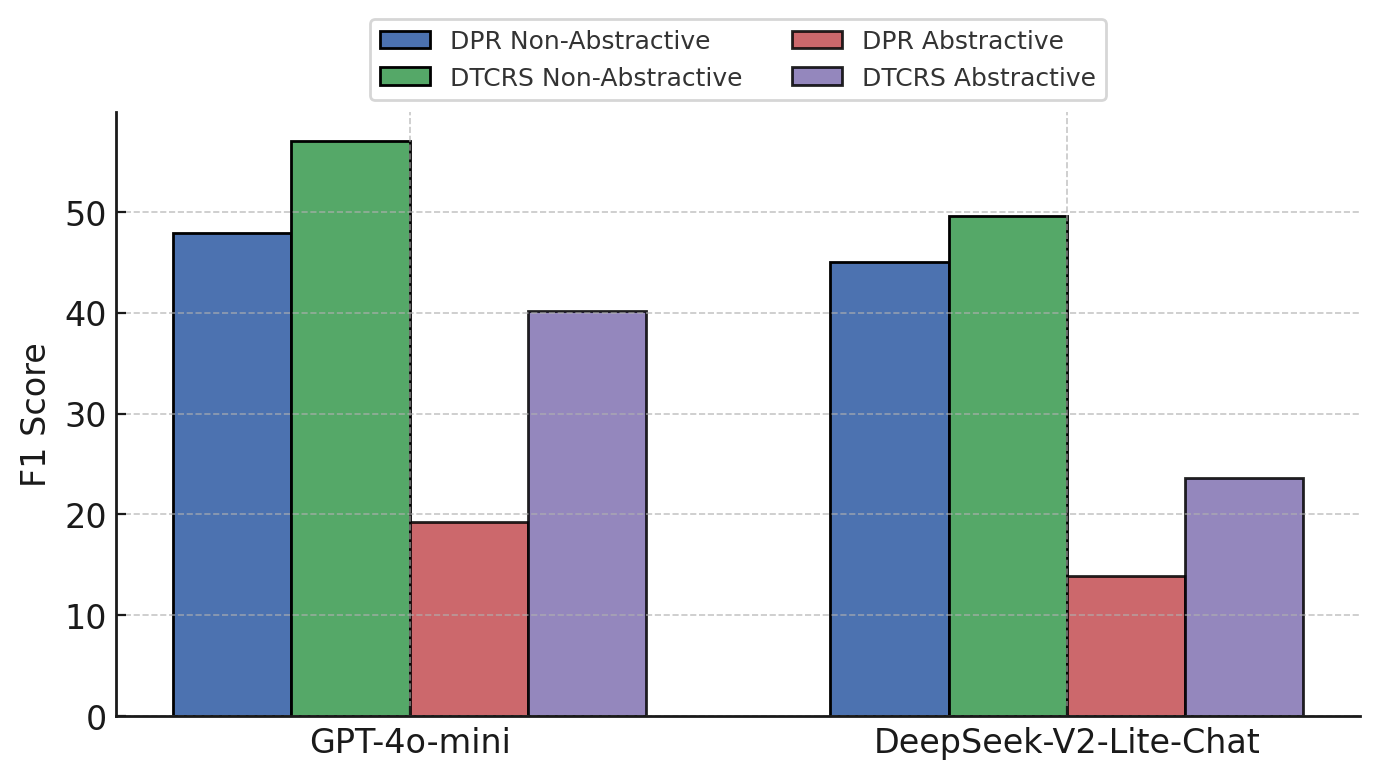}
    \caption{F1 scores comparison of DTCRS and DPR on abstractive vs. non-abstractive questions.}
    \label{fig:abs_vs_nonabs}
\end{figure}

We further compare performance on abstractive and non-abstractive questions. As shown in Figure~\ref{fig:abs_vs_nonabs}, DTCRS consistently outperforms DPR on both types, with larger gains for abstractive questions. These results highlight the particular advantage of DTCRS on tasks that require multi-hop reasoning and information synthesis.

\subsection{Ablation Study}
\begin{table*}[h]
\centering
\resizebox{\textwidth}{!}{%
\begin{tabular}{lcccccccc}
\toprule
\multirow{3}{*}{\makecell[c]{\textbf{Model}}} & \multicolumn{2}{c}{\textbf{Total}} & \multicolumn{2}{c}{\textbf{Extractive}} & \multicolumn{2}{c}{\textbf{Abstractive}} & \multicolumn{2}{c}{\textbf{Boolean}} \\
\cmidrule(lr){2-3} \cmidrule(lr){4-5} \cmidrule(lr){6-7} \cmidrule(lr){8-9}
 & GPT-4o-mini & \makecell{DeepSeek-V2- \\ Lite-Chat} & GPT-4o-mini & \makecell{DeepSeek-V2- \\ Lite-Chat} & GPT-4o-mini & \makecell{DeepSeek-V2- \\ Lite-Chat} & GPT-4o-mini & \makecell{DeepSeek-V2- \\ Lite-Chat} \\
\midrule
DPR & 33.0 & 31.3 & 29.6 & 25.7 & 13.0 & 13.9 & 86.3 & 85.4 \\
w/o global & 43.8 & \textbf{38.6} & 40.4 & 32.4 & 23.5 & 21.0 & 86.2 & 86.2 \\
w/o classify & 37.2 & 33.7 & 28.0 & 24.0 & 22.9 & 20.2 & \textbf{90.1} & \textbf{87.4} \\
w/o contents & 44.1 & 38.1 & 40.5 & 32.1 & 22.7 & 20.6 & 87.3 & 84.1 \\
\textbf{DTCRS} & \textbf{44.5} & 38.3 & \textbf{41.3} & \textbf{32.5} & \textbf{23.6} & \textbf{21.1} & 88.2 & 85.3 \\
\bottomrule
\end{tabular}
}
\caption{Ablation study results on QASPER using GPT-4o-mini and DeepSeek-V2-Lite-Chat, showing the F1 scores of each component on different question types.}
\label{tab:ablation_qasper}
\end{table*}

We perform ablation experiments using GPT-4o-mini and Deepseek-v2-Lite-Chat on the QASPER dataset to examine the contribution of each module to performance improvement. Specifically, we consider removing (1) global clustering and replacing it with hierarchical clustering, (2) the question type classifier, and (3) the Table of Contents (ToC) generator. The more complete ablation results will be presented in Appendix~\ref{sec:appendix2}.

As shown in Table~\ref{tab:ablation_qasper}, we analyze the performance of different DTCRS modules on various question types and draw the following conclusions: Removing global clustering has no significant impact on the results but improves clustering efficiency. We will analyze the efficiency of global and hierarchical clustering in the next section. Removing the ToC generator leads to a performance drop, indicating that it helps combine sub-questions with query semantics and document structure, improving the relevance between summaries and the question. The most notable impact is from removing the question classifier. This is because for extractive questions, DPR performs better, and introducing recursive summarization in such cases can lead to performance degradation. However, for abstractive questions, recursive summarization is beneficial. Since recursive summarization is not suitable for all types of questions, it is crucial to decide whether to incorporate it based on the question type.

\begin{table}[h]
\centering
\resizebox{\columnwidth}{!}{%
\begin{tabular}{lcc}
\toprule
\textbf{Variant} & \textbf{GPT-4o-mini} & \textbf{DeepSeek-V2-Lite-Chat} \\
\midrule
DPR & 33.0 & 31.3 \\
w/o global & 43.8 & \textbf{38.6} \\ 
w/o classifier & 37.2 & 33.7 \\
w/o ToC & 44.1 & 38.1 \\
w/o question decomposer & 37.5 & 32.9 \\
DTCRS & \textbf{44.5} & 38.3 \\
\bottomrule
\end{tabular}%
}
\caption{Ablation Study on Total Questions (F1 Scores)}
\label{tab:decomp_ablation}
\end{table}

\subsection{Analysis}
\begin{figure}[t]
  \centering
  \includegraphics[scale=0.3]{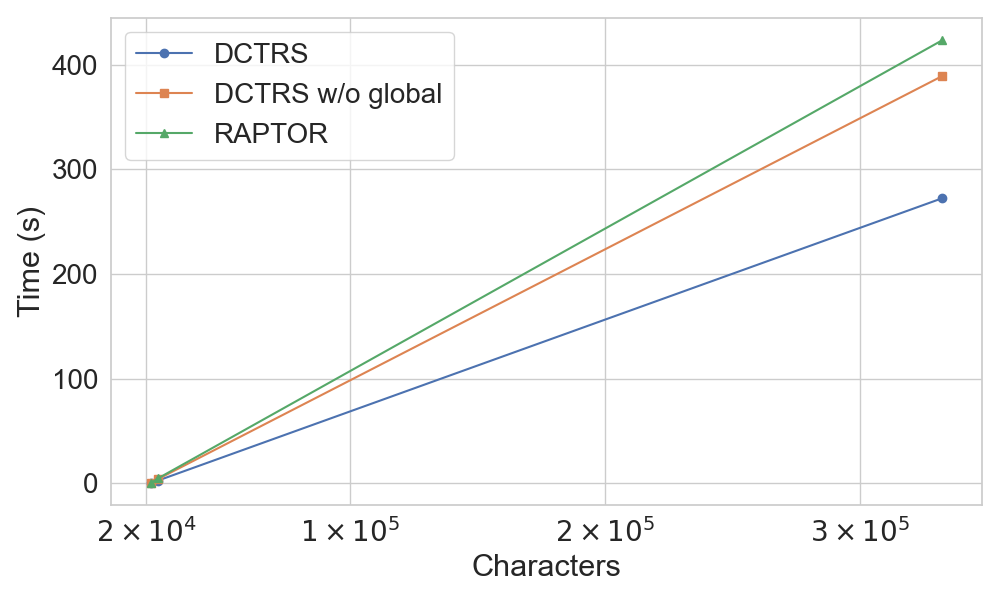}
  \caption{The sample outputs of RAPTOR and DTCRS: the yellow-highlighted parts in the evidence indicate information relevant to the question, the orange-highlighted parts represent redundant evidence included by RAPTOR compared to DTCRS, and the red bold text in the answer denotes the ground truth.}
  \label{fig:1}
\end{figure}

\begin{table}[h]
\centering
\resizebox{\columnwidth}{!}{%
    \begin{tabular}{lcccc}
        \toprule
        \multirow{2}{*}{\makecell[c]{\textbf{Layer}}} 
        & \multicolumn{2}{c}{\textbf{Average Number of Nodes}} 
        & \multicolumn{2}{c}{\textbf{Evidence Node Coverage}} \\
        \cmidrule(lr){2-3} \cmidrule(lr){4-5}
        & \textbf{DTCRS} & \textbf{RAPTOR} & \textbf{DTCRS} & \textbf{RAPTOR} \\
        \midrule
        0 (leaf) & 252  & 252  & 75.39\% & 70.83\% \\
        1        & 3.96 & 54   & 19.84\% & 20.83\% \\
        2        & 1.03 & 10   & 3.96\%  & 8.33\%  \\
        \bottomrule
    \end{tabular}%
}
\caption{Comparison of node statistics across layers. Evidence node coverage is the ratio of the number of evidence nodes in the layer to the total number of evidence nodes.}
\label{tab:node_stats}
\end{table}
\begin{figure}[t]
  \centering
  \includegraphics[scale=0.4]{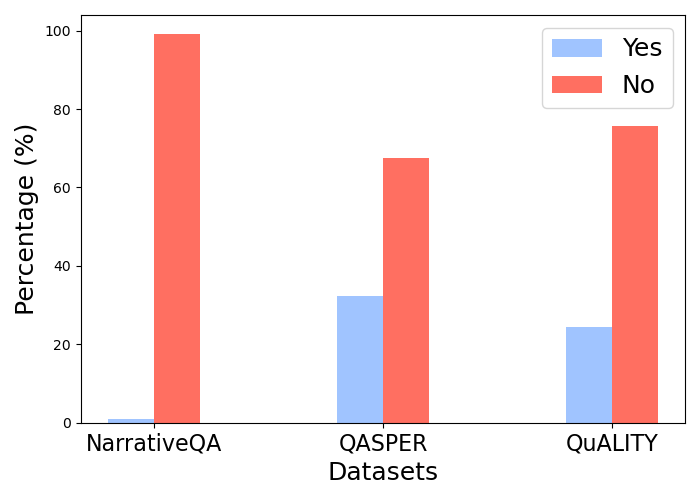} 
  \caption{Classification results using GPT-4o-mini as the classifier on different datasets: ``Yes'' indicates abstractive questions, ``No'' indicates non-abstractive questions.}
  \label{fig:3}
\end{figure}

\begin{figure}[t]
  \includegraphics[width=\columnwidth]{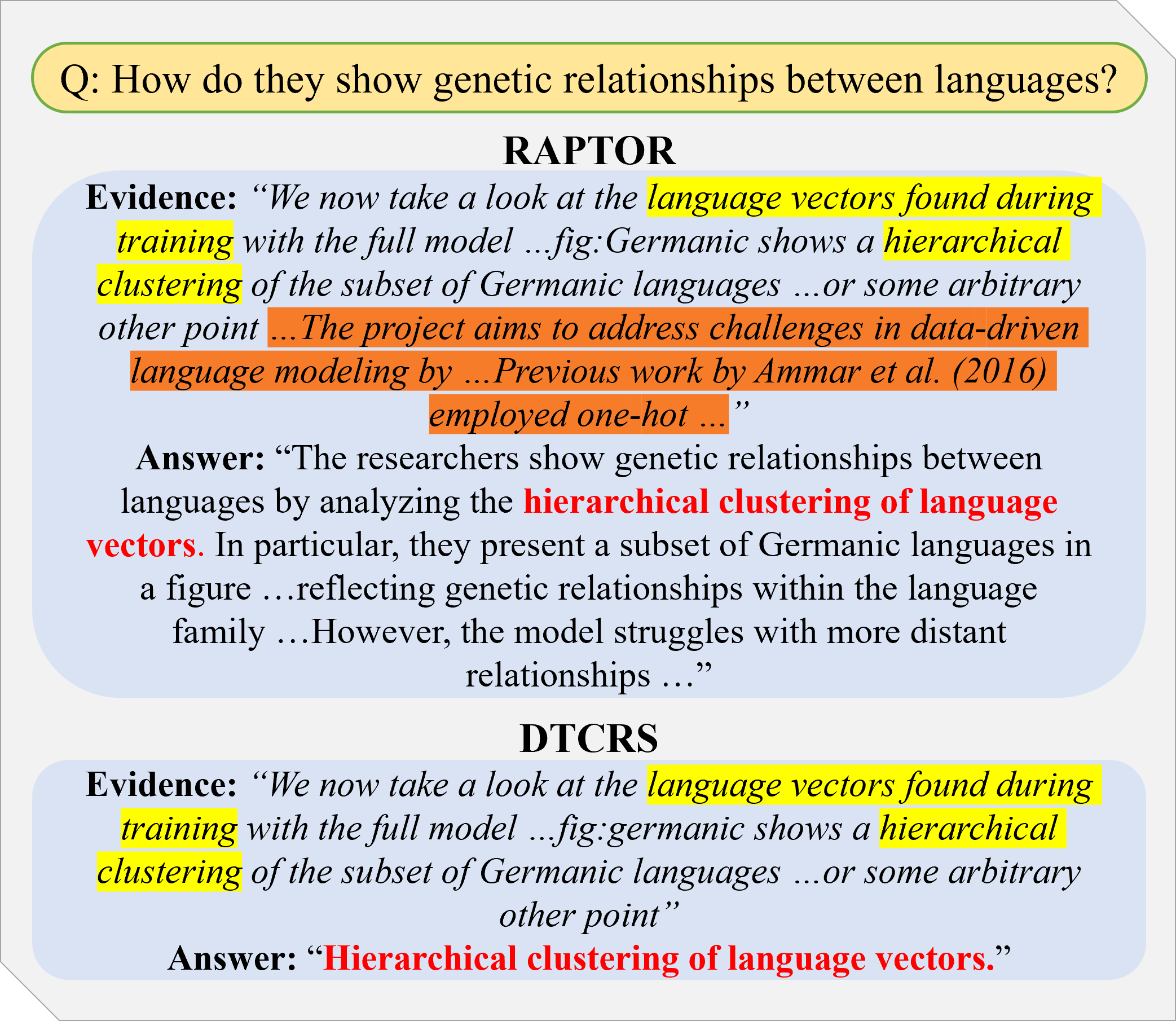}
  \caption{The sample outputs of RAPTOR and DTCRS: the yellow-highlighted parts in the evidence indicate information relevant to the question, the orange-highlighted parts represent redundant evidence included by RAPTOR compared to DTCRS, and the red bold text in the answer denotes the ground truth.}
  \label{fig:case}
\end{figure}

To evaluate whether our method reduces the summary tree construction time, we examine the relationship between document length and summary tree construction time, as shown in Figure~\ref{fig:1}. Overall, all methods exhibit a linear trend. Compared to DTCRS without global clustering and RAPTOR, DTCRS shows increasing time efficiency benefits as document length grows, indicating that our approach effectively reduces summary tree construction time by decreasing the number of clusters and incorporating global clustering.

Furthermore, we analyze the distribution of nodes across different layers to investigate whether DTCRS addresses the issue of redundant summary nodes. As shown in Table~\ref{tab:node_stats}, we report the average number of nodes and the evidence node coverage rate for the first three layers. DTCRS has significantly fewer summary nodes in the first and second layers than RAPTOR. Meanwhile, the evidence node coverage rate in the leaf layer and the first layer, which play a crucial role in answer generation, remains comparable to RAPTOR. This suggests that DTCRS effectively reduces the number of summary nodes while preserving the query-relevant information, thereby addressing the redundancy issue in summary nodes. The sample outputs in Figure~\ref{fig:case} further support this conclusion. Although both DTCRS and RAPTOR can retrieve evidence relevant to the question, DTCRS provides more concise evidence, which helps prevent the generation of irrelevant text in the answer.

In Section \ref{sec:main_results}, we find that the experimental results of DTCRS on NarrativeQA are not ideal. We believe this is because NarrativeQA primarily consists of simple extractive questions and lacks abstractive questions. To validate our point, we analyze the classification results of the question classifier on different datasets. As shown in Figure~\ref{fig:3}, the proportion of abstractive questions in QASPER and QuALITY exceeds 20\%, which leads to a significant performance improvement for DTCRS compared to RAPTOR and DPR. The QuALITY dataset benefits even more due to its higher proportion of abstractive questions. In contrast, NarrativeQA has almost no abstractive questions, so the performance of DTCRS is limited.

\paragraph{Computational Efficiency.}
Although our method requires several LLM calls, the main bottleneck is in clustering and summarization. As shown in Table~\ref{tab:overhead}, DTCRS reduces summary-layer nodes by 92.2\%, and the summary tree construction time (excluding preprocessing) is reduced by 80.95\% compared to RAPTOR.

\begin{table}[h]
\centering
\footnotesize
\begin{tabular}{lcc}
\toprule
\textbf{Method} & \textbf{Nodes} & \textbf{Time (s)} \\
\midrule
DTCRS  & 4.99  & 40.82 \\
RAPTOR & 64.00 & 214.39 \\
\bottomrule
\end{tabular}
\caption{Summary layer node count and construction time (excluding preprocessing) for DTCRS and RAPTOR.}
\label{tab:overhead}
\end{table}

\section{Conclusion}
We propose DTCRS, which dynamically generates a summary tree based on document structure and query semantics. First, it generates a table of contents for the document and then decomposes complex questions into simpler sub-questions based on the table of contents. The embeddings of these sub-questions serve as the initial cluster centers. Through recursive clustering and text summarization, DTCRS constructs a hierarchical summary tree. DTCRS achieves significant performance improvements on three QA datasets while requiring less time for construction. We conduct experiments to analyze DTCRS’s performance on different types of questions, providing insights for future research on the applicability of recursive summarization.

\section*{Ethics Statement}
We follow the new ACL Policy on AI Writing Assistance and use AI purely for language assistance in the paper. After careful consideration, we believe that our paper complies with the ACL ethics policy and does not introduce any additional ethical concerns.

\section*{Limitations}
The question classifier may make incorrect classifications, leading to the use of inappropriate retrieval methods, which can degrade performance. For long documents that exceed the model's input limit, the table of contents generated by the ToC generator may be incomplete, limiting the ability to summarize the document information effectively. In addition, due to constraints in computational resources and time, we only conduct experiments using the larger-scale GPT-4 model on the QASPER dataset, while smaller language models are used for the other datasets. Further evaluation with larger models on a broader range of datasets is needed in future work.

\section*{Acknowledgments}
We would like to thank the anonymous reviewers for their insightful and constructive feedback, which helped us improve the clarity, completeness, and overall quality of our work.
\bibliography{custom}

\appendix

\section{Prompt}
\label{sec:appendix1}
\begin{figure*}[h!]
    \centering
    \includegraphics[width=0.9\textwidth]{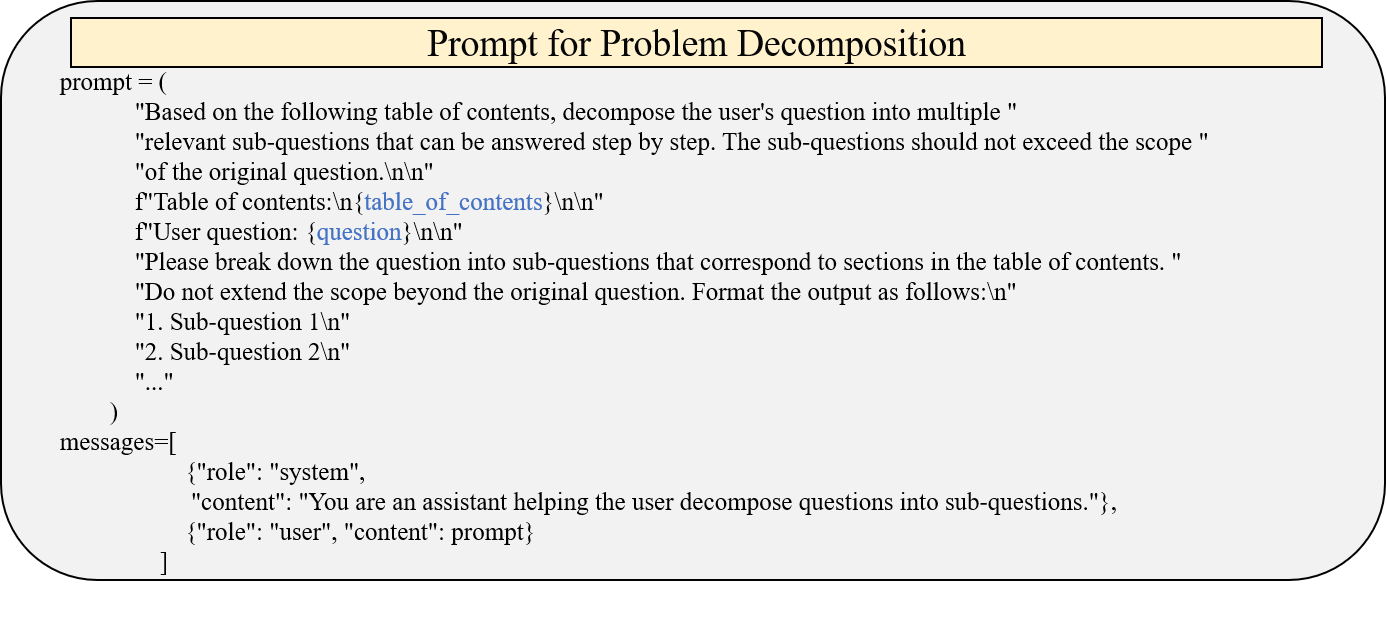}
    \caption{Prompt for Problem Decomposition}
    \label{fig:prompt1}
\end{figure*}

\begin{figure*}[h!]
    \centering
    \includegraphics[width=0.9\textwidth]{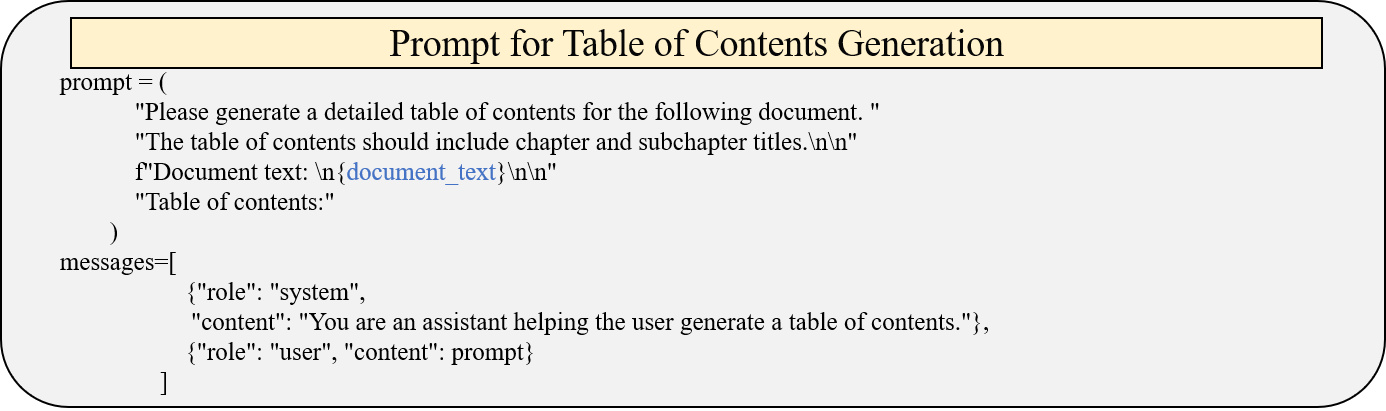}
    \caption{Prompt for Table of Contents Generation}
    \label{fig:prompt2}
\end{figure*}

\begin{figure*}[h!]
    \centering
    \includegraphics[width=0.9\textwidth]{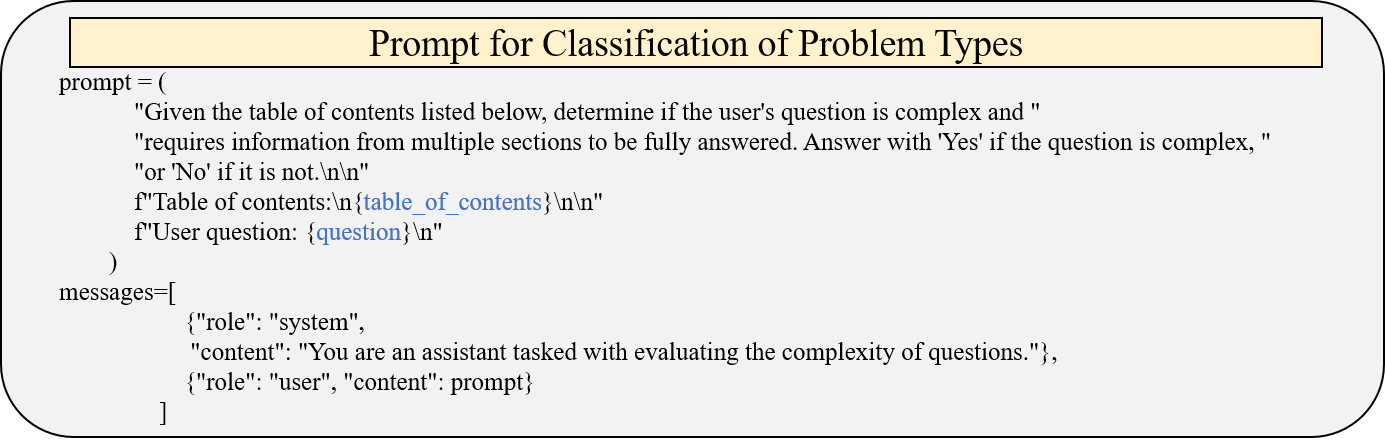}
    \caption{Prompt for Classification of Problem Types}
    \label{fig:prompt3}
\end{figure*}

\begin{figure*}[h!]
    \centering
    \includegraphics[width=0.9\textwidth]{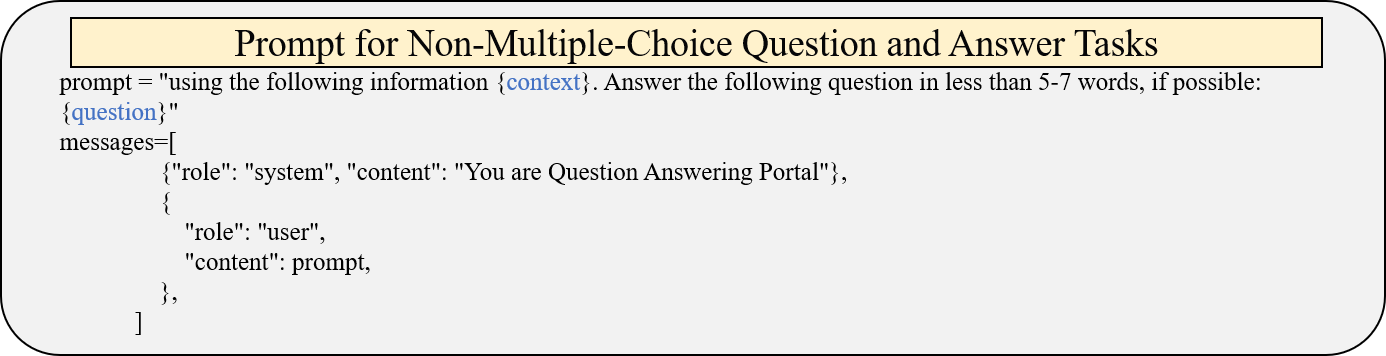}
    \caption{Prompt for Non-Multiple-Choice Question and Answer Tasks}
    \label{fig:prompt4}
\end{figure*}

\begin{figure*}[h!]
    \centering
    \includegraphics[width=0.9\textwidth]{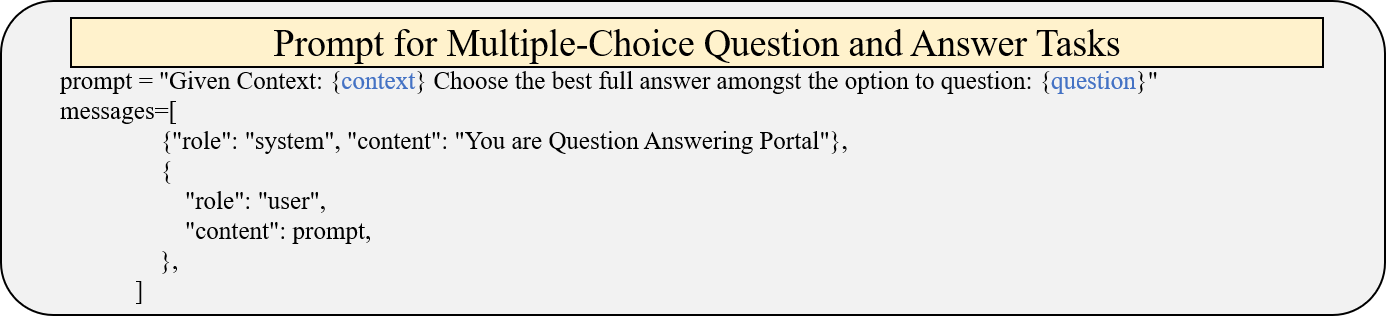}
    \caption{Prompt for Multiple-Choice Question and Answer Tasks}
    \label{fig:prompt5}
\end{figure*}

\begin{figure*}[h!]
    \centering
    \includegraphics[width=0.9\textwidth]{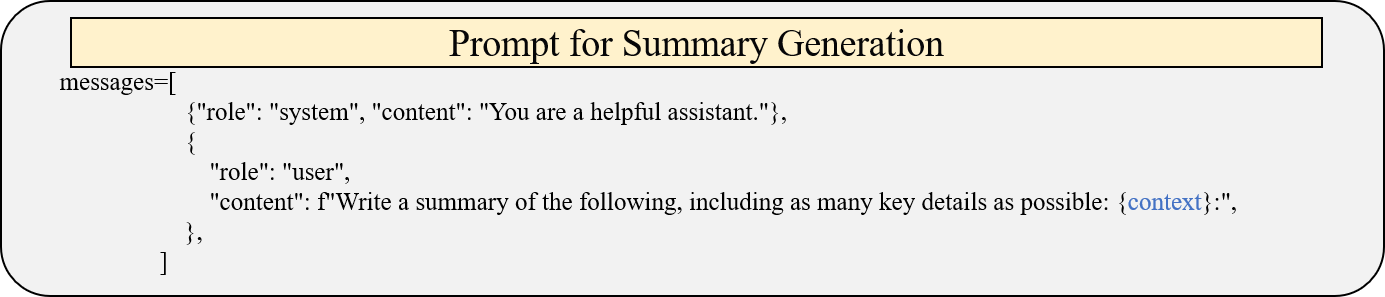}
    \caption{Prompt for Summary Generation}
    \label{fig:prompt6}
\end{figure*}

The prompts used in this study, depicted in Figures \ref{fig:prompt1} to \ref{fig:prompt6}, are designed to guide the model through a range of specific tasks effectively. 
\begin{itemize}
    \item \textbf{Problem Decomposition:} Figure \ref{fig:prompt1} shows a prompt that decomposes the user’s question into sub-questions based on the table of contents.

    \item \textbf{Table of Contents Generation:} Figure \ref{fig:prompt2} prompts the model to create a structured table of contents for the document.

    \item \textbf{Classification of Problem Types:} Figure \ref{fig:prompt3} valuate whether a user's question is complex and requires information from multiple sections of a document to be fully answered.

    \item \textbf{Non-Multiple-Choice Question and Answer:} Figure \ref{fig:prompt4} answers questions concisely (e.g., QASPER, NarrativeQA) using the provided context.

    \item \textbf{Multiple-Choice Question and Answer:} Figure \ref{fig:prompt5} selects the best answer from multiple options based on context (e.g., QuALITY).

    \item \textbf{Summary Generation:} Figure \ref{fig:prompt6} generates a concise summary of the given context, capturing key details.
\end{itemize}

\section{Complete Experimental Results}
\label{sec:appendix2}

Tables ~\ref{tab:results} and~\ref{tab:metric_comparison} show the complete experimental results compared with DPR and RAPTOR.

\begin{table*}[htbp]
\centering
\resizebox{\textwidth}{!}{%
\begin{tabular}{lcccccccc}
\toprule
\multirow{3}{*}{\makecell[c]{\textbf{Model}}} & \multicolumn{2}{c}{\textbf{Total}} & \multicolumn{2}{c}{\textbf{Extractive}} & \multicolumn{2}{c}{\textbf{Abstractive}} & \multicolumn{2}{c}{\textbf{Boolean}} \\
\cmidrule(lr){2-3} \cmidrule(lr){4-5} \cmidrule(lr){6-7} \cmidrule(lr){8-9}
 & GPT-4o-mini & \makecell{DeepSeek-V2- \\ Lite-Chat} & GPT-4o-mini & \makecell{DeepSeek-V2- \\ Lite-Chat} & GPT-4o-mini & \makecell{DeepSeek-V2- \\ Lite-Chat} & GPT-4o-mini & \makecell{DeepSeek-V2- \\ Lite-Chat} \\
\midrule
DPR & 33.0 & 31.3 & 29.6 & 25.7 & 13.0 & 13.9 & 86.3 & 85.4 \\
RAPTOR & 34.5 & 34.6 & 24.4 & 23.6 & 17.2 & 17.7 & 87.2 & 86.8 \\
w/o global & 43.8 & 38.6 & 40.4 & 32.4 & 23.5 & 21.0 & 86.2 & 86.2 \\
w/o classify & 37.2 & 33.7 & 28.0 & 24.0 & 22.9 & 20.2 & 90.1 & 87.4 \\
w/o contents & 44.1 & 38.1 & 40.5 & 32.1 & 22.7 & 20.6 & 87.3 & 84.1 \\
\textbf{DTCRS} & \textbf{44.5} & \textbf{38.3} & \textbf{41.3} & \textbf{32.5} & \textbf{23.6} & \textbf{21.1} & \textbf{88.2} & \textbf{85.3} \\
\bottomrule
\end{tabular}
}

\caption{Complete results on QASPER using GPT-4o-mini and DeepSeek-V2-Lite-Chat.}
\label{tab:results}
\end{table*}
\begin{table*}[htbp]
\centering
\resizebox{\textwidth}{!}{%
\begin{tabular}{lcccccccc}
\toprule
\multirow{3}{*}{\makecell[c]{\textbf{Model}}} & \multicolumn{2}{c}{\textbf{ROUGE-L}} & \multicolumn{2}{c}{\textbf{BLEU-1}} & \multicolumn{2}{c}{\textbf{BLEU-4}} & \multicolumn{2}{c}{\textbf{METEOR}} \\
\cmidrule(lr){2-3} \cmidrule(lr){4-5} \cmidrule(lr){6-7} \cmidrule(lr){8-9}
 & GPT-4o-mini & \makecell{DeepSeek-V2- \\ Lite-Chat} & GPT-4o-mini & \makecell{DeepSeek-V2- \\ Lite-Chat} & GPT-4o-mini & \makecell{DeepSeek-V2- \\ Lite-Chat} & GPT-4o-mini & \makecell{DeepSeek-V2- \\ Lite-Chat} \\
\midrule
DPR & 26.5 & 21.3 & 21.4 & 15.2 & 1.3 & 0.5 & 14.4 & 15.0 \\
RAPTOR & 25.0 & 25.4 & 21.2 & 17.9 & 1.1 & 1.0 & 14.1 & 18.3 \\
w/o global & \textbf{26.9} & 27.9 & 21.5 & 18.9 & 1.3 & \textbf{1.3} & 14.7 & 19.7 \\
w/o classify & 26.2 & 25.3 & 21.4 & 17.3 & 1.2 & 0.7 & \textbf{15.1} & 17.7 \\
w/o contents & 26.7 & 21.4 & 21.4 & 15.2 & 1.1 & 0.6 & 14.3 & 15.2 \\
\textbf{DTCRS} & 26.0 & \textbf{28.1} & 21.1 & \textbf{19.1} & \textbf{1.3} & 0.9 & 14.2 & \textbf{20.1} \\
\bottomrule
\end{tabular}
}
\caption{Complete results on NarrativeQA using GPT-4o-mini and DeepSeek-V2-Lite-Chat.}
\label{tab:metric_comparison}
\end{table*}

\section{Sample Output}
\label{sec:appendix3}
We present a detailed sample output in Table~\ref{tab:case results}.
\FloatBarrier
\begin{table*}[htbp]
\centering
\resizebox{1\textwidth}{!}{
\begin{tabular}{|p{0.2\textwidth}|p{0.35\textwidth}|p{0.35\textwidth}|}
\hline
\textbf{Question} & \multicolumn{2}{p{0.7\textwidth}|}{How do they show genetic relationships between languages?} \\
\hline
\textbf{Reference Answer} & \multicolumn{2}{p{0.7\textwidth}|}{By doing hierarchical clustering of word vectors} \\
\hline
\textbf{Method} & \textbf{Predicted Answer} & \textbf{Predicted Evidence} \\
\hline
\textbf{RAPTOR} & 
\begin{minipage}[t]{0.35\textwidth}
The researchers show genetic relationships between languages by analyzing the hierarchical clustering of language vectors. In particular, they present a subset of Germanic languages in a figure (fig:germanic), where the clustering closely mirrors the established genetic relationships within the Germanic language family. This suggests that the language vectors, while not explicitly designed to detect genealogical relationships, capture meaningful linguistic similarities, reflecting genetic relationships within the language family. However, the model struggles with more distant relationships, such as connecting Germanic languages to Celtic languages, indicating that the model is more successful at capturing closer, more evident linguistic relationships.

Therefore, the genetic relationships between languages are demonstrated through clustering, where the model's learned language vectors show a strong correlation with known language family trees, like those of the Germanic languages.
\end{minipage} &
\begin{minipage}[t]{0.35\textwidth}
However, our purpose is not to detect genealogical relationships... The only supervision... is a language identifier as a one-hot encoding

\textbf{In this paper, we examine a language model that utilizes a predictive distribution... Our focus is on investigating the implications of moving beyond the training corpus's language points}

By means of cross-entropy... Figure FIGREF12 plots the interpolation results...

\textbf{The project aims to address challenges in data-driven language modeling... Previous work by Ammar et al. (2016) employed one-hot}

The research focuses on developing... \textbf{distinguishing it from previous studies, such as Johnson et al. (2016)}...
\end{minipage} \\
\hline
\textbf{DTCRS} & 
\begin{minipage}[t]{0.35\textwidth}
Hierarchical clustering of language vectors.
\end{minipage} &
\begin{minipage}[t]{0.35\textwidth}
However, our purpose is not to detect genealogical relationships... The only supervision... is a language identifier as a one-hot encoding

\textbf{By means of cross-entropy... Figure FIGREF12 plots the interpolation results...}

In the figure English is (weakly) grouped with the West Germanic languages... Generating Text: Since our language model is conditioned on a language vector...
\end{minipage} \\
\hline
\end{tabular}
}
\caption{Comparison of RAPTOR and DTCRS Results.}
\label{tab:case results}
\end{table*}

\clearpage
\section{Inference Parameters}
\label{sec:appendix4}
Table~\ref{tab:temperature_settings} shows the temperatures for all steps that require reasoning with an LLM.
\begin{table}[H]
\centering
\begin{tabular}{lc}
\toprule
\textbf{Process Step} & \textbf{Temperature} \\
\midrule
Question Type Classification & 0 \\
Table of Contents Generation & 0 \\
Question Decomposition & 0 \\
Question Answering & 0 \\
Summary Generation & 0.3 \\
\bottomrule
\end{tabular}
\caption{Temperature settings for different process steps.}
\label{tab:temperature_settings}
\end{table}

\newpage

\section{UMAP Parameter Settings and Effects}
\label{sec:appendix5}

Key UMAP parameters include \texttt{n\_neighbors}, \texttt{n\_components}, and \texttt{metric}, which control the number of neighbors for manifold learning, the target dimension for reduction, and the distance metric, respectively. In our main experiments, we set \texttt{n\_neighbors} to 10, \texttt{metric} to cosine, and determine \texttt{n\_components} as:
\[
\resizebox{0.9\linewidth}{!}{%
  $\texttt{n\_components}
   =\min\bigl(\texttt{dim},\,\texttt{len(embeddings)} - 2\bigr)$%
}
\]
where \texttt{dim} is set to 10, and \texttt{len(embeddings)} denotes the number of text chunk embeddings.

We further evaluated the impact of different UMAP parameters on QASPER, as shown below:

\begin{table}[h]
\centering
\footnotesize
\begin{tabular}{cccc}
\toprule
\textbf{n\_neighbors} & \textbf{dim} & \textbf{Predicted F1 Score} \\
\midrule
10   & 10  & 58.5 \\
50   & 10  & 56.3 \\
50   & 50  & 60.0 \\
100  & 100 & 60.8 \\
\bottomrule
\end{tabular}
\caption{Effects of different UMAP parameters on QASPER performance.}
\label{tab:appendix_umap}
\end{table}

\section{Evaluation Script}
\label{sec:appendix6}
We use the NLTK library \citep{bird2006nltk} to compute ROUGE-L, BLEU, and METEOR scores, where the evaluation parameters for the ROUGE-L score are shown in Table~\ref{tab:params}. For questions with multiple reference answers, we select the one with the highest score.

\begin{table}[H]
\centering
\begin{tabular}{lc}
\toprule
\textbf{Parameter} & \textbf{Value} \\
\midrule
max\_n & 4 \\
limit\_length & True \\
length\_limit & 100 \\
length\_limit\_type & words \\
apply\_avg & True \\
apply\_best & True \\
alpha & 0.5 \\
weight\_factor & 1.2 \\
stemming & True \\
\bottomrule
\end{tabular}
\caption{Parameter settings for ROUGE-L.}
\label{tab:params}
\end{table}

\end{document}